\documentclass{article}

\usepackage{arxiv}

\usepackage[utf8]{inputenc}
\usepackage[T1]{fontenc}   
\usepackage{hyperref}      
\usepackage{url}            
\usepackage{booktabs}       
\usepackage{amsfonts}       
\usepackage{nicefrac}       
\usepackage{microtype}      
\usepackage{graphicx}
\usepackage[numbers]{natbib}
\usepackage{doi}
\usepackage{amsmath}
\usepackage{multirow}
\usepackage{cleveref}
\usepackage{todonotes}
\usepackage{subcaption}
\usepackage{lscape} 
\usepackage{lipsum}
\usepackage{algorithm}
\usepackage[section]{placeins}
\usepackage{algpseudocode}  
\usepackage{float}
\usepackage{quantikz}  

\renewcommand{\headeright}{}
\renewcommand{\undertitle}{}

\title{Bridging quantum and classical computing for partial differential equations through multifidelity machine learning}

\author{
        \href{https://orcid.org/0009-0001-5361-3105}{\includegraphics[scale=0.06]{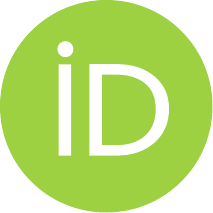}\hspace{1mm}Bruno Jacob}\\
	Pacific Northwest National Laboratory\\
	Richland, WA 99354 \\
	\texttt{bruno.jacob@pnnl.gov} \\
      \And  
        \href{https://orcid.org/0000-0002-6411-6198}{\includegraphics[scale=0.06]{Figures/orcid.pdf}\hspace{1mm}Amanda A. Howard}\\
	Pacific Northwest National Laboratory\\
	Richland, WA 99354 \\
	\texttt{amanda.howard@pnnl.gov} \\
     \And  
        \href{https://orcid.org/0000-0002-9928-5637}{\includegraphics[scale=0.06]{Figures/orcid.pdf}\hspace{1mm}Panos Stinis}\\
	Pacific Northwest National Laboratory\\
	Richland, WA 99354 \\
	\texttt{panagiotis.stinis@pnnl.gov} \\
}

\begin{document}

\maketitle

\begin{abstract}
Quantum algorithms for partial differential equations (PDEs) face severe practical constraints on near-term hardware: limited qubit counts restrict spatial resolution to coarse grids, while circuit depth limitations prevent accurate long-time integration. These hardware bottlenecks confine quantum PDE solvers to low-fidelity regimes despite their theoretical potential for computational speedup. We introduce a multifidelity learning framework that corrects coarse quantum solutions to high-fidelity accuracy using sparse classical training data, facilitating the path toward practical quantum utility for scientific computing. The approach trains a low-fidelity surrogate on abundant quantum solver outputs, then learns correction mappings through a multifidelity neural architecture that balances linear and nonlinear transformations. Demonstrated on benchmark nonlinear PDEs including viscous Burgers equation and incompressible Navier-Stokes flows via quantum lattice Boltzmann methods, the framework successfully corrects coarse quantum predictions and achieves temporal extrapolation well beyond the classical training window. This strategy illustrates how one can reduce expensive high-fidelity simulation requirements while producing predictions that are competitive with classical accuracy. By bridging the gap between hardware-limited quantum simulations and application requirements, this work establishes a pathway for extracting computational value from current quantum devices in real-world scientific applications, advancing both algorithm development and practical deployment of near-term quantum computing for computational physics.
\end{abstract}


\begin{figure}[t]
\centering
\includegraphics[width=\textwidth]{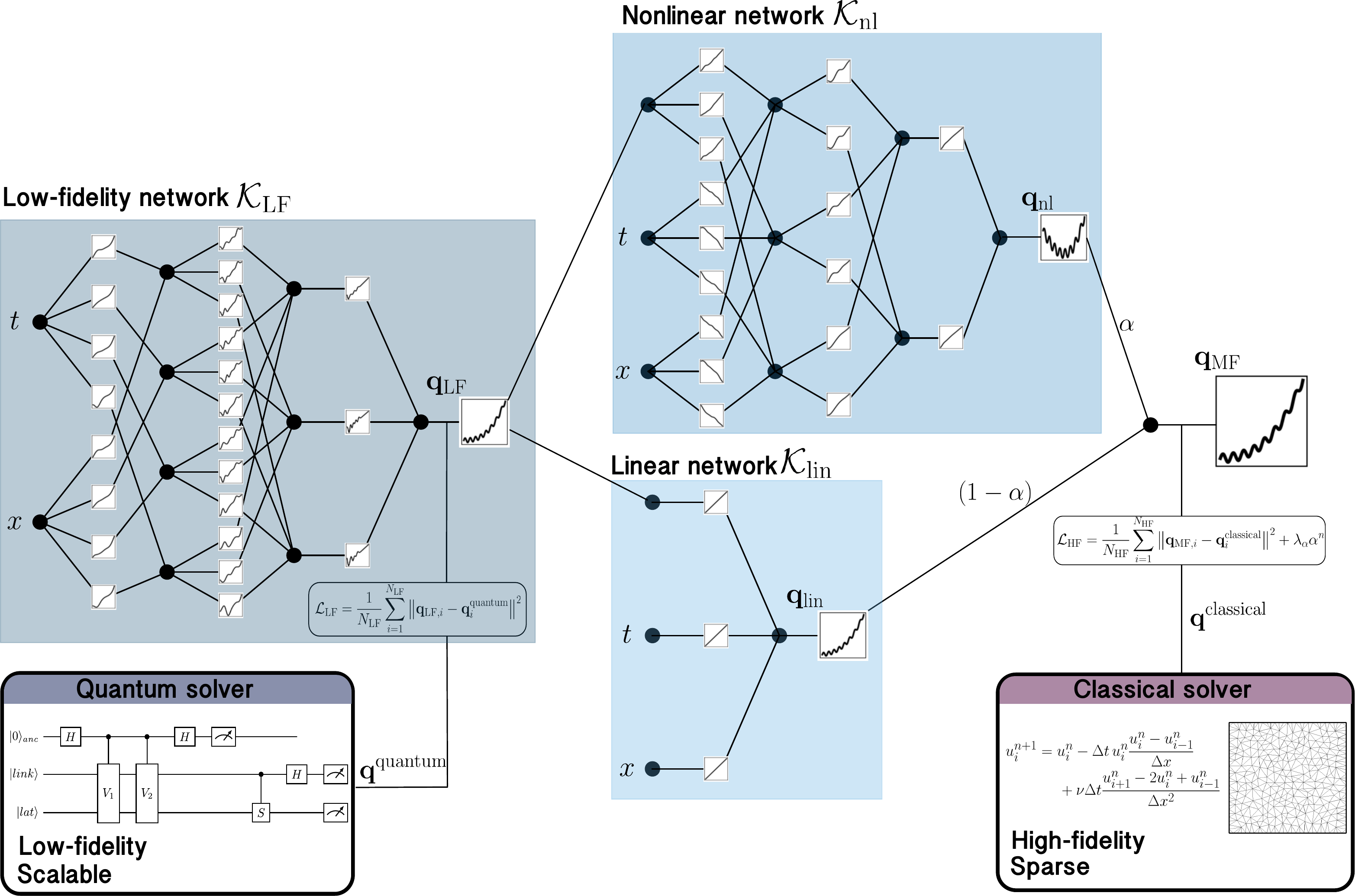}
\caption{Graphical abstract: overview of the hybrid quantum-classical multifidelity  
framework. The quantum solver (QLBM circuit, bottom left) generates abundant low-fidelity data $\mathbf{q}^{\text{quantum}}$ used to train the low-fidelity network $\mathcal{K}_{\text{LF}}$. The classical solver (bottom right) provides sparse high-fidelity data $\mathbf{q}^{\text{classical}}$ for training the correction networks. The nonlinear network $\mathcal{K}_{\text{nl}}$ and linear network $\mathcal{K}_{\text{lin}}$ take the low-fidelity predictions as input and are blended via the learned parameter $\alpha$ to produce the final multifidelity prediction $\mathbf{q}_{\text{MF}}$.}
\label{fig:hybrid_architecture}
\end{figure}

\section{Introduction}

The numerical solution of nonlinear partial differential equations (PDEs) remains a central computational challenge across scientific and engineering disciplines. High-resolution simulations of fluid dynamics, materials science phenomena, and reaction-diffusion systems often require prohibitive computational resources, particularly when fine spatial resolution and long-time integration are needed simultaneously. Classical numerical methods such as finite difference \cite{thomas2013numerical}, finite element \cite{reddy2005introduction}, and spectral techniques \cite{shen2011spectral} have reached maturity but face fundamental scalability limitations when problem complexity increases. These constraints motivate the exploration of alternative computational paradigms that can provide rapid approximate solutions, even at coarse resolution, which can then be systematically refined through data-driven techniques.

Quantum computing offers novel algorithmic approaches for scientific computing that exploit quantum mechanical principles such as superposition and entanglement~\cite{cao2019quantum}. Quantum algorithms for linear systems~\cite{harrow2009quantum} and physical simulations~\cite{lloyd1996universal} have demonstrated theoretical advantages, with recent work showing that quantum algorithms can solve certain classes of PDEs with solutions encoded in time polylogarithmic in the problem size~\cite{childs2021high}. Variational quantum algorithms have been developed to handle nonlinear PDEs by using multiple copies of quantum states to encode nonlinear terms~\cite{lubasch2020variational}. However, practical deployment for PDEs faces challenges including circuit depth limitations and noise in near-term quantum devices. Quantum lattice Boltzmann methods (QLBM) represent a particularly promising direction for fluid dynamics~\cite{succi2015quantum,mezzacapo2015quantum}, as they encode mesoscopic distribution functions directly on quantum registers and evolve them according to discrete kinetic theory. Recent work has demonstrated that such fully quantum lattice fluid algorithms can simulate canonical diffusion and Burgers equations on quantum simulators with good agreement with classical solutions~\cite{kocherla2023fully}. The QLBM-frugal algorithm introduced by Lee et al.~\cite{lee2024frugal} addresses resource constraints through a two-circuit architecture that separates collision and streaming operations, enabling implementation on near-term quantum hardware without requiring expensive subroutines like quantum amplitude estimation or matrix inversion.

Multifidelity methods combine datasets of varying accuracy and computational cost, leveraging correlations between them to achieve better predictions than either source alone~\cite{peherstorfer2018survey,fernandez2016review}. The central principle is training correction models that map inexpensive low-fidelity predictions toward high-fidelity targets using limited reference data~\cite{penwarden2022multifidelity}. Methodological approaches range from transfer learning~\cite{song2022transfer,chakraborty2021transfer,de2020transfer}, to feature-space fusion~\cite{guo2022multi,chen2023feature}, to composite neural network architectures~\cite{meng2020composite}. The composite approach separates the fidelity correction into additive linear and nonlinear branches, enabling robust training when high-fidelity samples are scarce. Extensions encompass physics-informed formulations~\cite{howard2023stacked}, continual learning~\cite{howard2024multifidelity}, operator networks~\cite{lu2022multifidelity,howard2023multifidelity,de2023bi,ahmed2023multifidelity}, and probabilistic variants~\cite{meng2021multi}, with applications in aerodynamics~\cite{zhang2021multi}, materials science~\cite{islam2021extraction}, and rheology~\cite{mahmoudabadbozchelou2021data,saadat2024data}.

The neural architecture employed for fidelity correction influences both accuracy and generalization. Kolmogorov-Arnold Networks (KANs)~\cite{liu2024kan} encode learnable univariate spline functions on edges rather than applying fixed nonlinearities at nodes, offering advantages in parameter efficiency and interpretability~\cite{shukla2024comprehensive}. The multifidelity KAN formulation~\cite{howard2025multifidelity} implements the composite correction within this spline-based framework, where spline order distinguishes linear from nonlinear correlation learning. This architecture enables extrapolation beyond the high-fidelity training window when additional low-fidelity data is available, reducing requirements for expensive reference simulations.

In this work, we integrate the QLBM-frugal quantum algorithm with multifidelity KANs to numerically solve nonlinear PDEs. The quantum solver provides abundant low-fidelity data on coarse grids constrained by near-term hardware, while the correction network learns to recover high-fidelity accuracy from sparse classical reference data. This framework applies broadly to problems where quantum algorithms offer fast approximate solutions.

The remainder of this paper is organized as follows. Section~\ref{sec:methods} describes the classical and quantum solvers, including detailed formulations of the QLBM-frugal algorithm and the multifidelity neural network architecture. Section~\ref{sec:results} presents numerical results for benchmark problems comparing quantum and classical approaches. Section~\ref{sec:conclusions} discusses implications and future directions.

\section{Methods}
\label{sec:methods}

\subsection{Problem Formulation}

We consider nonlinear time-dependent partial differential equations (PDEs) of the form:
\begin{equation}
\label{eq:general_pde}
\frac{\partial u}{\partial t} = \mathcal{N}_x(u; \boldsymbol{\lambda}), \quad x \in \Omega, \quad t \in [0, T]
\end{equation}
subject to boundary conditions
\begin{equation}
\label{eq:boundary}
\mathcal{B}_x(u) = 0, \quad x \in \partial\Omega
\end{equation}
and initial condition $u(x,0) = u_0(x)$. This general formulation encompasses a broad class of nonlinear evolution equations including nonlinear advection-diffusion equations, reaction-diffusion systems, and incompressible fluid flow when reformulated in appropriate variables. The solution field $u(x,t)$ may be scalar-valued (as in Burgers equation) or represent components of a vector field (as in the stream function-vorticity formulation of Navier-Stokes equations). The spatial operator $\mathcal{N}_x$ encodes the physical processes governing the system evolution and may include linear diffusion terms, nonlinear advection, reaction kinetics, or combinations thereof, parameterized by physical quantities $\boldsymbol{\lambda}$ such as viscosity, diffusion coefficients, or reaction rates. The spatial domain $\Omega \subset \mathbb{R}^d$ represents the physical region of interest with boundary $\partial\Omega$, and the boundary operator $\mathcal{B}_x$ enforces appropriate boundary conditions such as Dirichlet, Neumann, or periodic conditions depending on the physical problem.

\subsubsection{Spatial Discretization}

We discretize the spatial domain $\Omega$ using uniform grids with $N_x$ points, denoted $\{x_j\}_{j=0}^{N_x-1}$. The solution at time $t$ is represented as a finite-dimensional vector $\mathbf{u}(t) = [u_0(t), u_1(t), \ldots, u_{N_x-1}(t)]^T \in \mathbb{R}^{N_x}$ where each component $u_j(t) \approx u(x_j, t)$ approximates the continuous solution at grid point $x_j$. Spatial derivatives appearing in the operator $\mathcal{N}_x$ are approximated using standard numerical methods such as finite differences, spectral collocation, or finite elements. Following the method of lines approach, this spatial discretization transforms the continuous PDE~\eqref{eq:general_pde} into a semi-discrete system of ordinary differential equations in time:
\begin{equation}
\label{eq:semidiscrete_general}
\frac{d\mathbf{u}(t)}{dt} = \mathbf{f}(\mathbf{u}(t); \boldsymbol{\lambda}), \quad \mathbf{u}(0) = \mathbf{u}_0
\end{equation}
where $\mathbf{f}: \mathbb{R}^{N_x} \to \mathbb{R}^{N_x}$ is the discretized spatial operator encoding the physics. For nonlinear PDEs such as Burgers equation or Navier-Stokes, $\mathbf{f}$ is a nonlinear function of the state vector $\mathbf{u}$. The spatial discretization introduces truncation errors of order $\mathcal{O}(h^p)$ where $h$ is the grid spacing and $p$ is the method order. The resulting system of ODEs can then be integrated forward in time using standard time-stepping methods.

\subsection{High-fidelity Solvers}

We describe the classical high-fidelity numerical methods used to generate reference data for the 1D Burgers and 2D Navier-Stokes equations.

\subsubsection{Viscous Burgers Equation}
\label{sec:hf_burgers}

The 1D viscous Burgers equation is a prototypical nonlinear PDE combining unsteady advection and diffusion:
\begin{equation}
\frac{\partial u}{\partial t} + u\frac{\partial u}{\partial x} = \nu \frac{\partial^2 u}{\partial x^2}
\end{equation}
where $\nu$ is the kinematic viscosity. We discretize the spatial domain $x \in [0, L]$ using a uniform finite difference grid with periodic boundary conditions. The nonlinear advection term $u \partial_x u$ is discretized using a stable first-order upwind scheme, which adaptively selects forward or backward differences based on the local velocity sign to suppress non-physical oscillations near shocks. The diffusion term $\nu \partial_{xx} u$ is approximated using second-order central differences.

Time integration uses an explicit forward Euler method. The discretized update rule at grid point $i$ is:
\begin{equation}
u_i^{n+1} = u_i^n - \Delta t \left( u_i^n \frac{\delta_x u^n}{\delta x} \right) + \frac{\nu \Delta t}{\Delta x^2} \left( u_{i+1}^n - 2u_i^n + u_{i-1}^n \right)
\end{equation}
where $\frac{\delta_x u^n}{\delta x}$ denotes the upwind difference approximation. The timestep $\Delta t$ is chosen to satisfy stability constraints for the explicit scheme, including the CFL condition ($|u|_{\max} \Delta t / \Delta x \leq 1$) and the diffusion stability limit ($\nu \Delta t / \Delta x^2 \leq 0.5$). This classical solver provides high-resolution reference solutions on fine grids (e.g., $N_x=256$) to train the multifidelity correction models.

\subsubsection{Incompressible Navier-Stokes Equations}
\label{sec:hf_ns}

The incompressible Navier-Stokes equations govern viscous fluid flow:
\begin{align}
\frac{\partial u}{\partial t} + u\frac{\partial u}{\partial x} + v\frac{\partial u}{\partial y} &= -\frac{\partial p}{\partial x} + \frac{1}{\text{Re}}\left(\frac{\partial^2 u}{\partial x^2} + \frac{\partial^2 u}{\partial y^2}\right) \\
\frac{\partial v}{\partial t} + u\frac{\partial v}{\partial x} + v\frac{\partial v}{\partial y} &= -\frac{\partial p}{\partial y} + \frac{1}{\text{Re}}\left(\frac{\partial^2 v}{\partial x^2} + \frac{\partial^2 v}{\partial y^2}\right) \\
\frac{\partial u}{\partial x} + \frac{\partial v}{\partial y} &= 0
\end{align}
where $(u, v)$ are the velocity components, $p$ is the pressure, and $\text{Re} = UL/\nu$ is the Reynolds number controlling the relative importance of inertial to viscous forces.

For two-dimensional flows, we employ the stream function-vorticity formulation, which eliminates pressure and automatically satisfies the continuity constraint. The vorticity $\omega$ and stream function $\psi$ are defined by:
\begin{equation}
u = \frac{\partial \psi}{\partial y}, \quad v = -\frac{\partial \psi}{\partial x}, \quad \omega = \frac{\partial v}{\partial x} - \frac{\partial u}{\partial y}
\end{equation}
The governing equations become a coupled system:
\begin{align}
\frac{\partial \omega}{\partial t} + u\frac{\partial \omega}{\partial x} + v\frac{\partial \omega}{\partial y} &= \frac{1}{\text{Re}}\left(\frac{\partial^2 \omega}{\partial x^2} + \frac{\partial^2 \omega}{\partial y^2}\right) \label{eq:vorticity_transport} \\
\frac{\partial^2 \psi}{\partial x^2} + \frac{\partial^2 \psi}{\partial y^2} &= -\omega \label{eq:poisson_streamfunction}
\end{align}

The spatial domain is discretized using a uniform Cartesian grid with $N \times N$ points and spacing $h = 1/(N-1)$. Spatial derivatives are approximated using second-order central finite differences. The time integration proceeds through a fractional-step approach at each timestep $n$, advancing from time $t_n$ to $t_{n+1} = t_n + \Delta t$.

Given the current vorticity field $\omega^n$, we first solve the elliptic Poisson equation~\eqref{eq:poisson_streamfunction} for the stream function $\psi^n$ using an iterative Gauss-Seidel relaxation method:
\begin{equation}
\psi_{i,j}^{(m+1)} = \frac{1}{4}\left(\psi_{i+1,j}^{(m)} + \psi_{i-1,j}^{(m+1)} + \psi_{i,j+1}^{(m)} + \psi_{i,j-1}^{(m+1)} + h^2\omega_{i,j}^n\right)
\end{equation}
where superscript $m$ denotes the iteration counter. We iterate until the maximum residual falls below a specified tolerance. Velocity components are then computed from the converged stream function using centered finite differences:
\begin{equation}
u_{i,j} = \frac{\psi_{i,j+1} - \psi_{i,j-1}}{2h}, \quad v_{i,j} = -\frac{\psi_{i+1,j} - \psi_{i-1,j}}{2h}
\end{equation}
Finally, we advance the vorticity field forward in time using explicit forward Euler integration of the vorticity transport equation~\eqref{eq:vorticity_transport}, with both advective and diffusive terms discretized using second-order central finite differences. The timestep is chosen to satisfy stability constraints for the explicit scheme, including the CFL condition and the diffusion stability limit.

\subsection{Low-Fidelity Quantum Solvers}

We employ Quantum Lattice Boltzmann Methods (QLBM) as low-fidelity solvers. These solvers map the kinetic Boltzmann equation to quantum circuits, enabling the simulation of fluid dynamics on coarse grids suitable for near-term quantum hardware. We utilize the QLBM-frugal approach~\cite{lee2024frugal}, which emphasizes resource efficiency by splitting the coupled PDE system into distinct, optimized quantum circuits that can be executed concurrently.

\subsubsection{QLBM for Burgers Equation}

For the 1D viscous Burgers equation, we construct a one-dimensional analogue of the QLBM-frugal algorithm \cite{lee2024frugal} based on a D1Q3 lattice. At each lattice site $x_j$ and time $t_n$ we introduce distribution functions $f_k(x_j,t_n)$ associated with discrete velocities $e_k \in \{0,1,-1\}$ and weights $w_0 = 2/3$, $w_1 = w_2 = 1/6$. The macroscopic velocity is recovered by summing over the discrete velocities,
\begin{equation}
u(x_j,t_n) = \sum_{k=0}^{2} f_k(x_j,t_n).
\end{equation}
The nonlinear advection term $u \partial_x u$ is represented by treating Burgers equation as an advection-diffusion process with a state-dependent advection velocity $c_{\text{adv}}(x_j,t_n) = u(x_j,t_n)/2$, in close analogy with the advection-diffusion formulation used for QLBM in~\cite{lee2024frugal}. In lattice units (with $\Delta x = \Delta t = 1$), the local equilibrium distribution takes the form
\begin{equation}
\label{eq:burgers_eq_distribution}
f_k^{\text{eq}}(x_j,t_n)
= w_k\,u(x_j,t_n)\left(1 + \frac{e_k\,c_{\text{adv}}(x_j,t_n)}{c_s^2}\right),
\end{equation}
where $c_s^2 = 1/3$ is the lattice sound speed. The relaxation time $\tau$ controls the effective viscosity of the D1Q3 scheme. The corresponding classical lattice Boltzmann update is
\begin{equation}
\label{eq:burgers_lbm_update}
f_k(x_j + e_k, t_{n+1})
= f_k(x_j,t_n)
- \omega\left[f_k(x_j,t_n) - f_k^{\text{eq}}(x_j,t_n)\right],
\end{equation}
with relaxation rate $\omega = 1/\tau$. This update recovers a viscous Burgers equation for $u(x,t)$ in the hydrodynamic limit, with viscosity determined by the lattice parameters.

The quantum Burgers solver emulates the evolution~\eqref{eq:burgers_lbm_update} using a collision-streaming sequence on a joint lattice-link Hilbert space. The physical velocity profile $u(x_j,t_n)$ is first mapped to a normalized state vector whose amplitudes encode $u(x_j,t_n)$ on the lattice register, and this vector is replicated across the three link sectors corresponding to $e_0$, $e_1$, and $e_2$. The non-unitary BGK collision is implemented via block encoding of a diagonal collision matrix $A$ whose entries are the equilibrium coefficients
\begin{equation}
\label{eq:burgers_collision_diag}
a_{k,j} = w_k\left(1 + \frac{e_k\,c_{\text{adv}}(x_j,t_n)}{c_s^2}\right),
\end{equation}
combined with exponential relaxation in time. Specifically, we form a relaxed diagonal $D = (1-\omega)D_{\text{prev}} + \omega A$ and clip its entries so that $|D|<1$. Following~\cite{lee2024frugal}, this diagonal is embedded into a Linear Combination of Unitaries
\begin{equation}
\label{eq:burgers_lcu}
C_{1,2} = D \pm i\sqrt{I - D^2},
\end{equation}
so that $D = (C_1 + C_2)/2$ is realized as a sub-block of a larger unitary.

The quantum circuit introduces an ancilla qubit that controls the application of $C_1$ and $C_2$. Preparing the ancilla in $(|0\rangle + |1\rangle)/\sqrt{2}$, applying the controlled diagonals, and then applying a second Hadamard on the ancilla implements the collision operator in the ancilla-zero subspace, up to a known normalization factor. The streaming step follows collision and is implemented using controlled circular shift operators on the lattice register. The qubit register is partitioned into a link register $|link\rangle$ (encoding the discrete velocity direction) and a lattice register $|lat\rangle$ (encoding spatial position). Right and left shift unitaries act on $|lat\rangle$, controlled on link states corresponding to $e_1$ and $e_2$, thereby realizing the streaming $x_j \mapsto x_j \pm 1$ with periodic boundary conditions.

After collision and streaming, Hadamard gates are applied to the link register to sum over link directions and recover the macroscopic field. In the statevector simulation we directly read the amplitudes associated with the ancilla in $|0\rangle$ and the link register in the rest state, and analytically remove the normalization factors introduced by the block encoding and Hadamard operations. We additionally enforce conservation of total momentum $\sum_j u(x_j,t_n)$ at each time step by a scalar rescaling. This procedure yields a quantum D1Q3 evolution that is consistent with the classical lattice Boltzmann scheme~\eqref{eq:burgers_lbm_update} while remaining implementable on near-term quantum hardware.

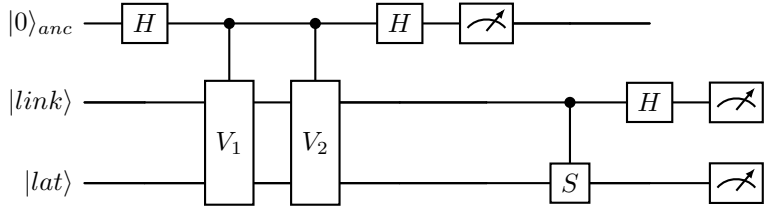
\begin{figure}[!htbp]
\centering
\begin{quantikz}
\lstick{$|0\rangle_{anc}$} & \gate{H} & \ctrl{1} & \ctrl{1} & \gate{H} & \meter{} & \qw & \qw \\
\lstick{$|link\rangle$} & \qw & \gate[2]{V_1} & \gate[2]{V_2} & \qw & \qw & \ctrl{1} & \gate{H} & \meter{} \\
\lstick{$|lat\rangle$} & \qw & \qw & \qw & \qw & \qw & \gate{S} & \qw & \meter{}
\end{quantikz}
\caption{Schematic quantum circuit for a single 1D Burgers QLBM step. The ancilla-based block encoding (LCU) implements the non-unitary collision operator via controlled diagonals $V_1$ and $V_2$, and the streaming operator $S$ acts on the lattice register $|lat\rangle$ controlled by the link register $|link\rangle$, realizing the shift $x_j \mapsto x_j \pm 1$. A final Hadamard on $|link\rangle$ represents summation over discrete velocities for macroscopic retrieval; in this work the circuit is classically emulated via statevector simulation rather than hardware measurements.}
\label{fig:burgers_circuit}
\end{figure}

\subsubsection{QLBM-Frugal for Navier-Stokes}

For 2D incompressible flows, we implement the QLBM-frugal algorithm~\cite{lee2024frugal} for the Navier-Stokes equations using the same stream function-vorticity formulation described in Section~\ref{sec:hf_ns}. QLBM-frugal exploits the decoupled structure of equations~\eqref{eq:vorticity_transport}--\eqref{eq:poisson_streamfunction} by utilizing two distinct, optimized quantum circuits: a \emph{vorticity circuit} and a \emph{stream function circuit}. This separation allows for concurrent execution on quantum hardware and significantly reduces the depth of individual circuits compared to unified solvers. In this work, We use a D2Q5 lattice with five velocity directions. The state encoding uses $n_{\text{spatial}} = 2\lceil \log_2 M \rceil$ qubits for the $M \times M$ spatial grid and $n_{\text{link}} = 3$ qubits for the velocity directions.

\paragraph{Vorticity Circuit:}
The vorticity transport equation behaves as an advection-diffusion system with a state-dependent advection velocity. The circuit evolves the vorticity distribution functions using a collision-streaming sequence. The collision operator is non-unitary and is implemented via block encoding of a diagonal matrix whose entries are constructed from local equilibrium coefficients based on the velocity field recovered from the stream function. A single ancilla qubit controls the Linear Combination of Unitaries (LCU) that realizes the BGK-relaxed collision operator, while the link register selects which shift operation is applied in the streaming step. In our implementation, vorticity boundary values are computed classically from Thom's formula and imposed after the quantum update rather than through an explicit boundary register.

\begin{figure}[!htbp]
\centering
\begin{quantikz}
\lstick{$|0\rangle_a$} & \gate{H} & \ctrl{1} & \ctrl{1} & \gate{H} & \qw & \qw & \qw \\
\lstick{$|link\rangle$} & \qw & \gate[2]{C_1} & \gate[2]{C_2} & \qw & \ctrl{1} & \gate{H} & \qw \\
\lstick{$|lat\rangle$} & \qw & \qw & \qw & \qw & \gate{S} & \qw & \qw
\end{quantikz}
\caption{Vorticity circuit for the D2Q5 QLBM-frugal scheme~\cite{lee2024frugal}. The ancilla qubit $|0\rangle_a$ controls block encoding of the non-unitary collision operator via diagonal gates $C_1$ (controlled on $|0\rangle$) and $C_2$ (controlled on $|1\rangle$). The streaming operator $S$ shifts the lattice register $|lat\rangle$ controlled by the link register $|link\rangle$. A final Hadamard layer on $|link\rangle$ sums over velocity directions for macroscopic retrieval. In the implementation, the circuit uses 12 qubits total (8 spatial + 3 link + 1 ancilla), and vorticity boundary conditions are imposed classically after each quantum update.}
\label{fig:vorticity_circuit}
\end{figure}
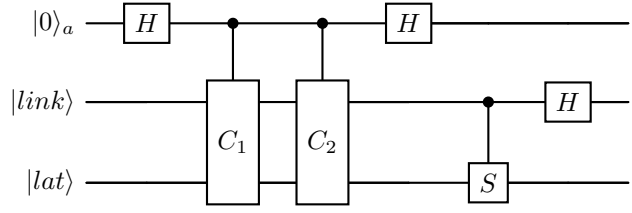

\paragraph{Stream Function Circuit:}
The Poisson equation~\eqref{eq:poisson_streamfunction} is solved using a similar lattice Boltzmann scheme but requires the addition of a source term $S = -\omega$. In our implementation the vorticity field is embedded into additional slots of the input state so that the block-encoded collision diagonal already incorporates the effect of the source term, again using an ancilla-controlled LCU construction. The subsequent streaming step acts on the spatial lattice in the D2Q5 stencil, and macroscopic stream function values are recovered by summing over link directions as in the advection-diffusion case, followed by classical enforcement of $\psi=0$ on the cavity walls.

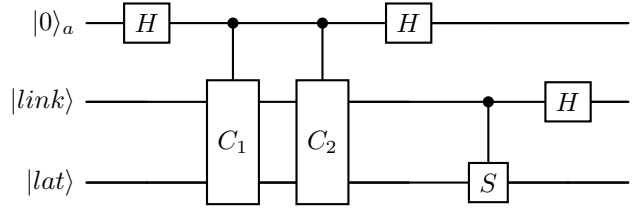
\begin{figure}[!htbp]
\centering
\begin{quantikz}
\lstick{$|0\rangle_a$} & \gate{H} & \ctrl{1} & \ctrl{1} & \gate{H} & \qw & \qw & \qw \\
\lstick{$|link\rangle$} & \qw & \gate[2]{C_1} & \gate[2]{C_2} & \qw & \ctrl{1} & \gate{H} & \qw \\
\lstick{$|lat\rangle$} & \qw & \qw & \qw & \qw & \gate{S} & \qw & \qw
\end{quantikz}
\caption{Stream function circuit for the D2Q5 QLBM-frugal scheme~\cite{lee2024frugal}. The structure mirrors the vorticity circuit: block-encoded collision via $C_1$ and $C_2$ controlled by ancilla $|0\rangle_a$, streaming $S$ controlled by $|link\rangle$, and Hadamard summation on $|link\rangle$ for macroscopic retrieval. The vorticity source term $S=-\omega$ is embedded directly into the input state amplitudes rather than via a separate source register. The circuit uses 13 qubits total (8 spatial + 3 link + 2 ancilla), and $\psi=0$ boundary conditions are imposed classically after each quantum update.}
\label{fig:stream_circuit}
\end{figure}

Both circuits utilize the "frugal" resource reduction strategy. Non-unitary collision operators, including BGK relaxation and boundary/source terms, are decomposed into a Linear Combination of Unitaries (LCU), $\mathcal{C} = (C_1 + C_2)/2$, requiring only one ancilla qubit and avoiding expensive coherent arithmetic for computing moments on the quantum device. Equilibrium distributions, boundary conditions, and source terms are computed classically between time steps and loaded into the diagonal collision gates ($C_1, C_2$). This hybrid approach leverages the classical computer for nonlinear algebra while reserving the quantum device for the parallel state evolution. Separating the vorticity and stream function solvers allows for concurrent execution and shallower individual circuits compared to a unified solver.

The complete algorithm alternates between these two circuits. At each time step $t$, the Stream Function Circuit computes $\psi^{t}$ given $\omega^{t-1}$. The classical co-processor then computes velocity fields from $\psi^{t}$ to construct the collision operator for the Vorticity Circuit, which subsequently computes $\omega^{t}$. This ping-pong execution updates the coupled system efficiently.

\subsection{Multifidelity Kolmogorov-Arnold Network Architecture}

We employ a multifidelity learning framework~\cite{howard2025multifidelity} to combine abundant low-fidelity quantum data with sparse high-fidelity classical data. The correction architecture leverages Kolmogorov-Arnold networks (KANs)~\cite{liu2024kan}, which differ from traditional multi-layer perceptrons by placing learnable activation functions on network edges rather than fixed activations on nodes. This design enables adaptive basis function learning that can capture complex functional relationships more efficiently.

In the KAN formulation, each edge activation $\varphi$ combines a fixed base function with an adaptive B-spline component to provide both stability and flexibility during training. The B-spline representation enables the network to learn smooth univariate transformations on each edge, with the overall multivariate function emerging from their composition across layers. This architecture offers advantages for our application: the adaptive activations can efficiently represent the correction mappings between quantum and classical solutions, while the spline-based parameterization naturally handles the continuous spatiotemporal inputs.

Let $\mathbf{q}$ denote the set of flow observables of interest (for example, velocity components $(u,v)$ in the cavity case and scalar velocity $u$ for the Burgers equation). The low-fidelity KAN $\mathcal{K}_{\text{LF}}$ maps spatiotemporal coordinates to these quantities:
\begin{equation}
\mathcal{K}_{\text{LF}}: (\mathbf{x}, t) \mapsto \mathbf{q}_{\text{LF}}(\mathbf{x},t),
\end{equation}
where $\mathbf{x}$ denotes spatial coordinates (scalar $x$ for 1D problems, vector $(x,y)$ for 2D). It is trained exclusively on abundant quantum solver data. The linear KAN $\mathcal{K}_{\text{lin}}$ takes as input both the spatiotemporal coordinates and the low-fidelity predictions:
\begin{equation}
\mathcal{K}_{\text{lin}}: (\mathbf{x}, t, \mathbf{q}_{\text{LF}}) \mapsto \mathbf{q}_{\text{lin}},
\end{equation}
and captures primarily linear correlations between low- and high-fidelity fields. The nonlinear KAN $\mathcal{K}_{\text{nl}}$ has the same input structure but employs higher-order B-splines:
\begin{equation}
\mathcal{K}_{\text{nl}}: (\mathbf{x}, t, \mathbf{q}_{\text{LF}}) \mapsto \mathbf{q}_{\text{nl}}.
\end{equation}

The final high-fidelity prediction is a convex combination of the linear and nonlinear outputs using a learnable parameter $\alpha \in [0,1]$:
\begin{equation}
\label{eq:multifidelity_blend}
\mathbf{q}_{\text{MF}} \coloneqq \mathcal{K}_{\text{HF}}(\mathbf{x},t) = \alpha \cdot \mathcal{K}_{\text{nl}}(\mathbf{x},t,\mathbf{q}_{\text{LF}}) + (1-\alpha) \cdot \mathcal{K}_{\text{lin}}(\mathbf{x},t,\mathbf{q}_{\text{LF}}).
\end{equation}
The parameter $\alpha$ is initialized to $0.5$ and learned during training. This formulation automatically determines whether linear or nonlinear correlations dominate, with regularization encouraging simpler linear models when they provide sufficient accuracy.

Training proceeds in two stages. First, the low-fidelity network $\mathcal{K}_{\text{LF}}$ is trained on the abundant quantum data to minimize the low-fidelity loss: 

\begin{equation}
\mathcal{L}_{\text{LF}} = \frac{1}{N_{\text{LF}}}\sum_{i=1}^{N_{\text{LF}}} \bigl\|\mathcal{K}_{\text{LF}}(\mathbf{x}_i,t_i) - \mathbf{q}_i^{\text{quantum}}\bigr\|^2,
\end{equation}
where $N_{\text{LF}}$ is the number of low-fidelity training samples spanning multiple temporal snapshots on the quantum grid. After freezing the low-fidelity network weights to preserve the learned approximation, we train the linear and nonlinear networks plus the blending parameter $\alpha$ on sparse high-fidelity data. The training objective is a composite loss function:

\begin{equation}
\label{eq:multifidelity_loss}
\mathcal{L}_{\text{HF}} = \frac{1}{N_{\text{HF}}}\sum_{i=1}^{N_{\text{HF}}} \bigl\|\mathcal{K}_{\text{HF}}(\mathbf{x}_i,t_i) - \mathbf{q}_i^{\text{classical}}\bigr\|^2 + \lambda_\alpha \alpha^n
\end{equation}
designed to balance prediction accuracy against model complexity. The first term is the mean squared error between multifidelity predictions $\mathcal{K}_{\text{HF}}$ and high-fidelity classical reference data $\mathbf{q}^{\text{classical}}$, where $N_{\text{HF}}$ is the number of high-fidelity training samples. This term drives the linear and nonlinear networks to learn the correlation between low- and high-fidelity data. The second term $\lambda_\alpha \alpha^n$ regularizes the blending parameter, with the exponent $n=4$, as adopted by \cite{howard2025multifidelity}, creating a strong preference for small $\alpha$ values and encouraging the framework to rely on the simpler linear network when it provides sufficient accuracy.

\section{Results}
\label{sec:results}

We validate the proposed quantum-classical hybrid framework on two benchmark problems representing different classes of nonlinear PDEs: the 1D viscous Burgers equation and 2D incompressible Navier-Stokes equations. Throughout this work, all quantum circuits are classically emulated using Qiskit's \cite{javadi2024quantum} statevector simulator. All experiments were conducted on an NVIDIA A100 Tensor Core GPU with 40GB of memory.

\subsection{Case Study 1: Viscous Burgers Equation}
\label{sec:burgers}

We consider the 1D viscous Burgers equation of Section~\ref{sec:methods} on the domain $x \in [0, 1]$ with periodic boundary conditions and a localized Gaussian pulse as initial condition:
\begin{equation}
u(x,0) = \tfrac{1}{2}\exp\bigl(-40(x-0.35)^2\bigr).
\end{equation}
We use viscosity $\nu=0.01$. This problem exhibits key challenges for PDE solvers due to the interplay of nonlinear advection and diffusion, which generates steep gradients and strong spatial variation. The multifidelity framework combines a quantum advection-diffusion solver providing coarse spatial resolution with a classical finite difference solver providing high spatial resolution.

We compare the classical finite difference solver of Section~\ref{sec:hf_burgers} with the quantum D1Q3 QLBM solver described above. The classical solver uses upwind advection and central difference diffusion with explicit forward Euler time integration on a fine grid ($N_x=256$), while the quantum solver applies the block-encoded collision-streaming operator on a coarse lattice ($N_x=16$) suitable for near-term devices. The effectiveness of the multifidelity framework is evaluated by comparing three levels of solution fidelity: raw quantum low-fidelity predictions, multifidelity corrected predictions, and classical high-fidelity reference solutions. The key challenge is accurately capturing wave propagation despite the 16$\times$ resolution gap between LF and HF grids; the correction networks must learn both spatial upsampling and systematic bias corrections in regions of steep gradients.

\subsubsection{Experimental Setup}

\begin{figure}[!htbp]
\centering
\includegraphics[width=0.6\textwidth]{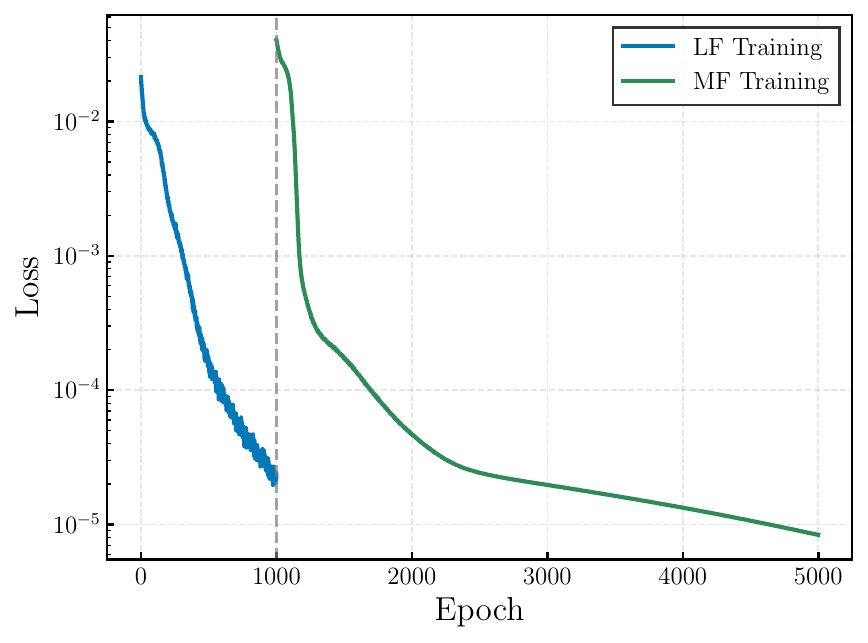}
\caption{Training loss evolution for the Burgers equation (B4). The vertical dashed line marks the transition from LF to MF training.}
\label{fig:burgers_loss}
\end{figure}

We conduct a systematic study of the multifidelity framework by varying two key hyperparameters: the blending regularization strength $\lambda_\alpha$ and the B-spline grid resolution $G$. The low-fidelity KAN uses layer dimensions $[2, 6, 6, 1]$ and the high-fidelity correction head uses $[3, 12, 12, 1]$. Training proceeds for 1000 LF epochs and 4000 HF epochs using Adam optimization with learning rate $10^{-3}$. The LF network is trained on quantum data spanning the full time interval $t \in [0, 0.5]$, while the HF correction networks are trained on classical data restricted to $t \leq 0.25$, requiring the model to extrapolate to later times.  Figure~\ref{fig:burgers_loss} shows the training loss evolution for the best-performing configuration (B4).

Table~\ref{tab:burgers_results} summarizes the results across all experimental configurations. We report relative $L^2$ errors on the training region ($t \leq 0.25$), extrapolation region ($t > 0.25$), and full domain, along with the learned blending parameter $\alpha$ and total training time.

\begin{table}[!htbp]
\centering
\caption{Multifidelity correction results for viscous Burgers equation. The baseline LF (quantum) solver achieves $L^2_{\text{full}} = 0.335$. Bold indicates best performance in each category.}
\label{tab:burgers_results}
\begin{tabular}{lccccccc}
\toprule
\textbf{Experiment} & $\lambda_\alpha$ & $G$ & $L^2_{\text{train}}$ & $L^2_{\text{extrap}}$ & $L^2_{\text{full}}$ & Final $\alpha$ & Time (s) \\
\midrule
\multicolumn{8}{l}{\textit{Effect of blending regularization $\lambda_\alpha$}} \\
B1: $\lambda_\alpha=0$ & $0$ & 5 & 0.0027 & 0.1159 & 0.0780 & 0.708 & 52.47 \\
B2: $\lambda_\alpha=10^{-6}$ & $10^{-6}$ & 5 & 0.0027 & 0.1154 & 0.0776 & 0.705 & 51.99 \\
B3: $\lambda_\alpha=10^{-5}$ & $10^{-5}$ & 5 & 0.0027 & 0.1112 & 0.0748 & 0.681 & 52.23 \\
B4: $\lambda_\alpha=10^{-4}$ & $10^{-4}$ & 5 & 0.0032 & \textbf{0.0914} & \textbf{0.0615} & 0.530 & 52.44 \\
B5: $\alpha=1$ (fixed) & $0$ & 5 & 0.0053 & 0.7173 & 0.4825 & 1.000 & 42.93 \\
\midrule
\multicolumn{8}{l}{\textit{Effect of B-spline grid resolution $G$}} \\
B6: $G=3$ & $0$ & 3 & 0.0072 & 0.1446 & 0.0974 & 0.739 & 49.69 \\
B1: $G=5$ (= B1 above) & $0$ & 5 & 0.0027 & 0.1159 & 0.0780 & 0.708 & 52.47 \\
B7: $G=7$ & $0$ & 7 & \textbf{0.0018} & 0.1911 & 0.1285 & 0.728 & 49.24 \\
\bottomrule
\end{tabular}
\end{table}

\subsubsection{Effect of Blending Regularization}

The blending parameter $\alpha$ controls the relative contribution of nonlinear versus linear corrections in \eqref{eq:multifidelity_blend}. The regularization term $\lambda_\alpha \alpha^4$ in the loss function~\eqref{eq:multifidelity_loss} penalizes large $\alpha$ values, encouraging the model to favor simpler linear corrections when possible.

Experiments B1--B5 reveal a clear trade-off between training accuracy and extrapolation generalization. Without regularization (B1, $\lambda_\alpha=0$), the model learns $\alpha \approx 0.71$, achieving low training error ($L^2=0.0027$) but moderate extrapolation error ($L^2=0.1159$). As regularization increases, the learned $\alpha$ decreases progressively: from $0.71$ (B1) to $0.53$ (B4, $\lambda_\alpha=10^{-4}$). This shift toward linear corrections yields slightly worse training error but substantially improved extrapolation performance; the extrapolation error drops from $0.1159$ to $0.0914$, an improvement of 21\%. At the opposite extreme, fixing $\alpha=1$ (B5, pure nonlinear correction) yields extrapolation error of $L^2_{\text{extrap}}=0.7173$, which is worse than the uncorrected LF solution, highlighting the importance of the linear network in the multifidelity architecture when used for temporal generalization.

The optimal configuration tested (B4, $\lambda_\alpha=10^{-4}$) achieves the best full-domain accuracy with $L^2_{\text{full}}=0.0615$, representing an 82\% reduction compared to the raw quantum LF solution ($L^2=0.335$). This result demonstrates that moderate regularization produces corrections that generalize better beyond the training window.

\subsubsection{Effect of B-Spline Grid Resolution}

The B-spline grid parameter $G$ controls the expressiveness of the KAN activation functions. Larger $G$ enables finer-grained learned activations at the cost of increasing the number of parameters.

Experiments B1, B6, and B7 compare $G \in \{3, 5, 7\}$ with no blending regularization. The results reveal a bias-variance trade-off: $G=7$ achieves the best training fit ($L^2_{\text{train}}=0.0018$) but the worst extrapolation ($L^2_{\text{extrap}}=0.1911$), indicating overfitting to the training window. The coarser grid $G=3$ shows the opposite behavior with higher training error ($L^2_{\text{train}}=0.0072$) but improved extrapolation ($L^2_{\text{extrap}}=0.1446$). The intermediate choice $G=5$ achieves the best extrapolation performance ($L^2_{\text{extrap}}=0.1159$) while maintaining low training error, providing the optimal balance between expressiveness and generalization.

\begin{figure}[!htbp]
	\centering
	\includegraphics[width=0.95\textwidth]{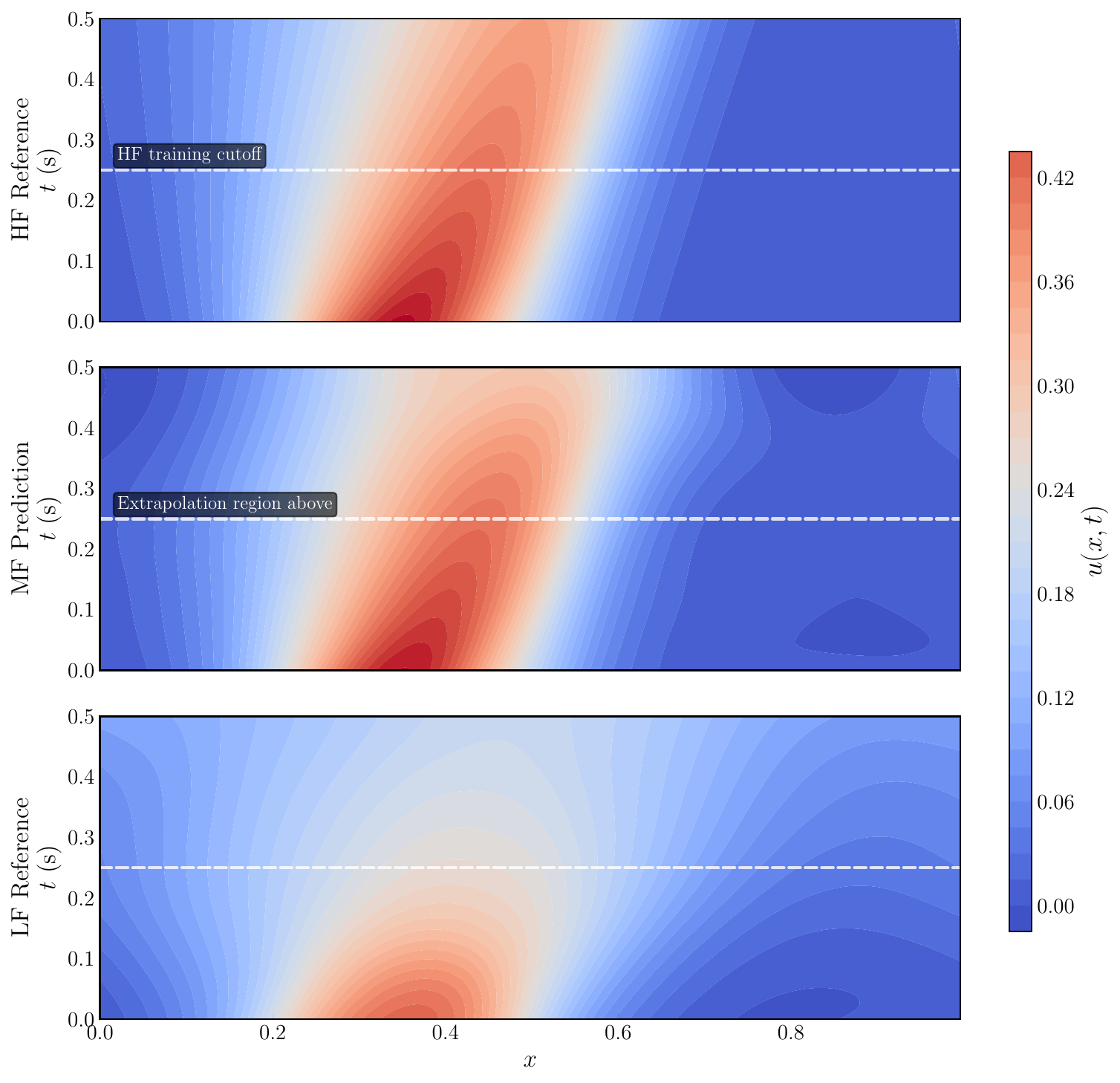}
	\caption{Spacetime evolution of velocity $u(x,t)$ for the viscous Burgers equation. Top: high-fidelity classical reference. Middle: multifidelity prediction from the best performing configuration (B4). Bottom: low-fidelity quantum (QLBM) solution. The dashed line at $t=0.25$ marks the HF training cutoff. In the extrapolation region ($t > 0.25$), the MF predictions are generated using only the LF quantum data and the trained correction model; no HF data is used. The HF reference is shown solely for validation.}
	\label{fig:burgers_spacetime}
	\end{figure}

Figure~\ref{fig:burgers_spacetime} visualizes the velocity field evolution for the best-performing configuration (B4). The HF reference (top) shows the characteristic nonlinear wave steepening and rightward propagation of the initial Gaussian pulse. The MF prediction (middle) closely matches the reference throughout both the training region ($t \leq 0.25$) and the extrapolation region ($t > 0.25$). The LF quantum solution (bottom) captures the qualitative dynamics but exhibits excessive numerical diffusion and spatial smoothing due to the coarse $16$-point grid.

\begin{figure}[!htbp]
\centering
\includegraphics[width=0.9\textwidth]{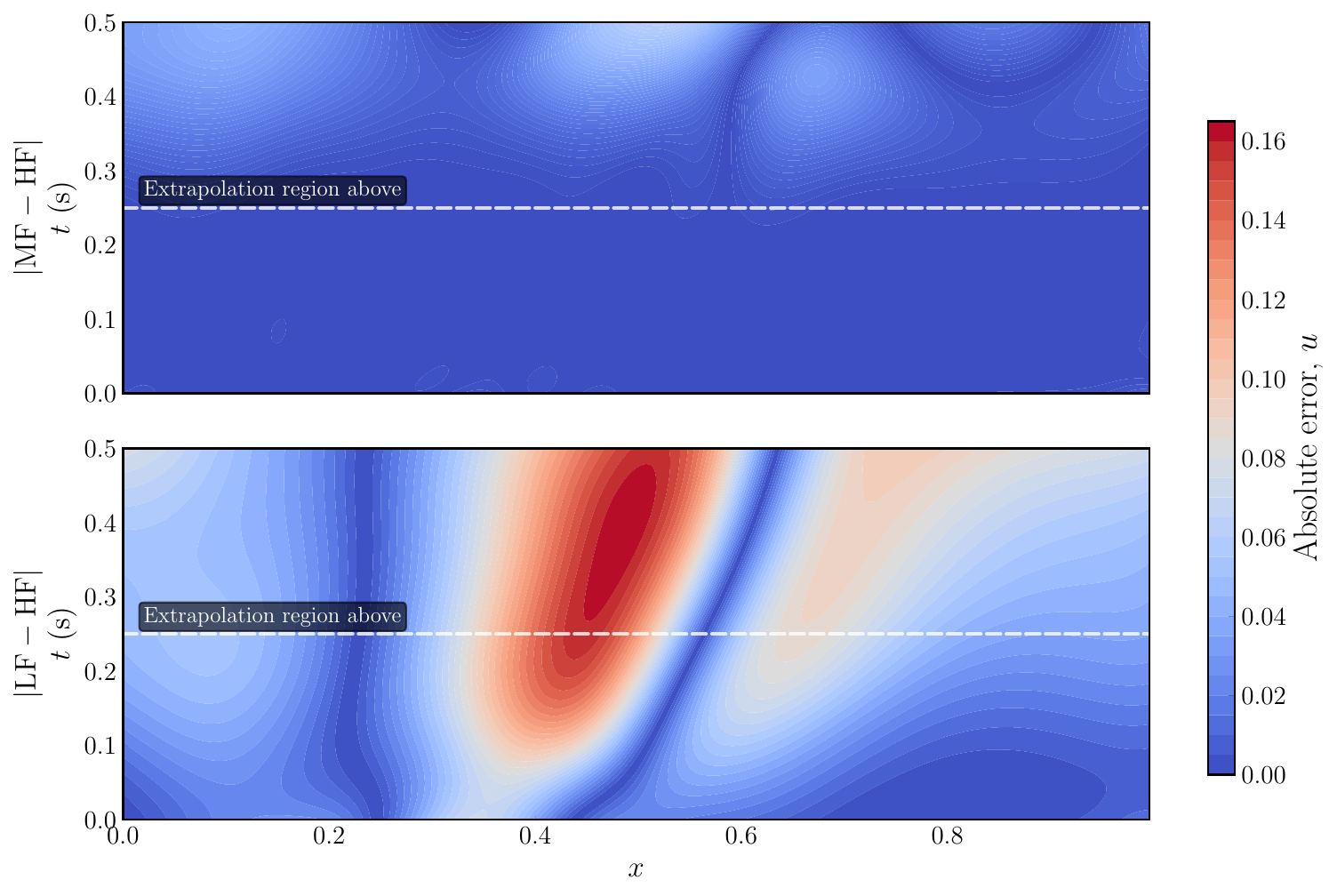}
\caption{Pointwise absolute errors for the viscous Burgers equation (B4). Top: multifidelity error $|u_{\text{MF}} - u_{\text{HF}}|$. Bottom: low-fidelity error $|u_{\text{LF}} - u_{\text{HF}}|$. The MF correction reduces maximum errors from $\sim 0.14$ to $\sim 0.02$, with particularly strong improvement in the wavefront region where the quantum solver exhibits the largest discrepancies.}
\label{fig:burgers_error}
\end{figure}

Figure~\ref{fig:burgers_error} compares the pointwise absolute errors $|u_{\text{MF}} - u_{\text{HF}}|$ and $|u_{\text{LF}} - u_{\text{HF}}|$. The MF correction reduces errors by an order of magnitude across the entire spatiotemporal domain. The LF error concentrates along the propagating wavefront where nonlinear steepening creates steep gradients that the coarse quantum grid cannot resolve. The MF correction successfully learns to compensate for these systematic biases, maintaining low errors even in the extrapolation region where no HF training data was provided.

\subsection{Case Study 2: 2D Lid-Driven Cavity Flow}
\label{sec:cavity}

We consider the lid-driven cavity problem using the stream function-vorticity formulation of Section~\ref{sec:hf_ns} on the domain $(x,y) \in [0,1]^2$. The problem consists of a square cavity with a moving lid at the top boundary ($u=1$, $v=0$ at $y=1$) and stationary no-slip walls on all other boundaries. The initial condition is quiescent flow ($\omega = \psi = 0$). We use Reynolds number $\text{Re} = 100$. This configuration exhibits key challenges for PDE solvers due to the interaction between the shear-driven primary vortex and corner singularities at the lid-wall junctions. The multifidelity framework combines a quantum QLBM solver providing coarse spatial resolution with a classical finite difference solver providing high spatial resolution.

We compare the classical finite difference solver of Section~\ref{sec:hf_ns} with the quantum D2Q5 QLBM-frugal solver described above. The classical solver uses the stream function-vorticity formulation on a fine $64 \times 64$ grid, while the quantum solver applies the two-circuit collision-streaming operator on a coarse $16 \times 16$ lattice suitable for near-term devices. The key challenge is accurately capturing the recirculating flow structure and boundary layer dynamics despite the $4\times$ resolution gap between LF and HF grids.

\subsubsection{Experimental Setup}

\begin{figure}[!htbp]
\centering
\includegraphics[width=0.6\textwidth]{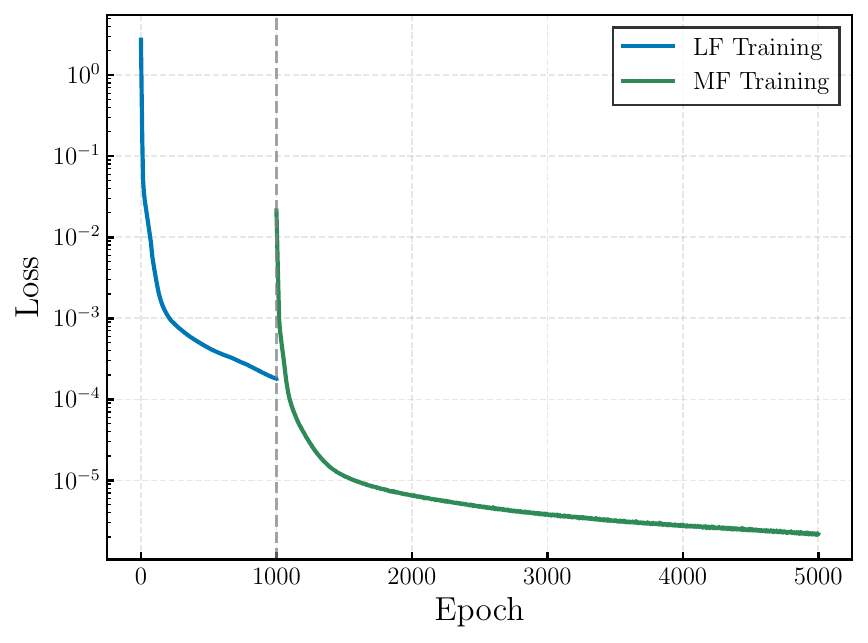}
\caption{Training loss evolution for the lid-driven cavity (C3). The vertical dashed line marks the transition from LF to MF training.}
\label{fig:cavity_loss}
\end{figure}

We conduct a systematic study of the multifidelity framework by varying two key aspects: the blending regularization strength $\lambda_\alpha$ and the network width. The low-fidelity KAN uses layer dimensions $[3, W_{\text{LF}}, W_{\text{LF}}, 2]$ and the high-fidelity correction head uses $[5, W_{\text{HF}}, W_{\text{HF}}, 2]$, where the baseline configuration uses $W_{\text{LF}}=8$ and $W_{\text{HF}}=10$. All experiments use B-spline grid resolution $G=5$. Training proceeds for 1000 LF epochs and 4000 HF epochs using Adam optimization with learning rate $10^{-3}$. The LF network is trained on quantum data spanning the full time interval $t \in [0, 3.0]$, while the HF correction networks are trained on classical data restricted to $t \leq 2.0$, requiring the model to extrapolate to later times. Figure~\ref{fig:cavity_loss} shows the training loss evolution for the best-performing configuration (C3).

Table~\ref{tab:cavity_results} summarizes the results across all experimental configurations. We report relative $L^2$ errors for both velocity components on the training region ($t \leq 2.0$), extrapolation region ($t > 2.0$), and full domain, along with the learned blending parameter $\alpha$ and total training time.

\begin{table}[!htbp]
\centering
\caption{Multifidelity correction results for lid-driven cavity flow at Re=100. The baseline LF (quantum) solver achieves relative $L^2$ error $L^2_{\text{full},u} = 0.285$ and $L^2_{\text{full},v} = 0.412$. Bold indicates best performance in each category.}
\label{tab:cavity_results}
\begin{tabular}{lccccccc}
\toprule
\textbf{Experiment} & $\lambda_\alpha$ & $L^2_{\text{train},u}$ & $L^2_{\text{extrap},u}$ & $L^2_{\text{full},u}$ & $L^2_{\text{full},v}$ & Final $\alpha$ & Time (s) \\
\midrule
\multicolumn{8}{l}{\textit{Effect of blending regularization $\lambda_\alpha$}} \\
C1: $\lambda_\alpha=10^{-3}$ & $10^{-3}$ & 0.009 & 0.141 & 0.114 & 0.203 & 0.105 & 297.03 \\
C2: $\lambda_\alpha=10^{-6}$ & $10^{-6}$ & 0.007 & 0.130 & 0.105 & 0.171 & 0.502 & 298.66 \\
C3: $\lambda_\alpha=10^{-5}$ & $10^{-5}$ & 0.008 & \textbf{0.097} & \textbf{0.079} & 0.167 & 0.299 & 308.26 \\
C4: $\lambda_\alpha=10^{-4}$ & $10^{-4}$ & 0.008 & 0.137 & 0.111 & \textbf{0.150} & 0.175 & 310.16 \\
C5: $\alpha=1$ (fixed) & $0$ & 0.007 & 0.143 & 0.116 & 0.162 & 1.000 & 305.65 \\
\midrule
\multicolumn{8}{l}{\textit{Effect of network width ($\lambda_\alpha=10^{-5}$, LF/HF layer dimensions; C3 uses $[3,8,8,2]$/$[5,10,10,2]$)}} \\
C6: $[3,5,5,2]$/$[5,5,5,2]$ & $10^{-5}$ & 0.024 & 0.182 & 0.148 & 0.267 & 0.349 & 252.17 \\
C7: $[3,10,10,2]$/$[5,10,10,2]$ & $10^{-5}$ & 0.008 & 0.185 & 0.150 & 0.202 & 0.303 & 308.11 \\
C8: $[3,15,15,2]$/$[5,15,15,2]$ & $10^{-5}$ & \textbf{0.004} & 0.166 & 0.134 & 0.152 & 0.263 & 415.61 \\
\bottomrule
\end{tabular}
\end{table}

\subsubsection{Effect of Blending Regularization}

Experiments C1--C5 examine the role of the blending parameter $\alpha$ and its regularization. As with the Burgers case, the results reveal a trade-off between training accuracy and extrapolation generalization, though the optimal regularization strength differs.

With strong regularization (C1, $\lambda_\alpha=10^{-3}$), the learned $\alpha$ converges to approximately $0.1$, favoring predominantly linear corrections. As regularization weakens, $\alpha$ increases progressively, reaching $0.5$ for the weakest regularization (C2). The optimal configuration is C3 with $\lambda_\alpha=10^{-5}$, which learns $\alpha \approx 0.3$ and achieves the best horizontal velocity accuracy with $L^2_{\text{full},u}=0.079$, representing a 72\% reduction compared to the raw quantum LF solution. This moderate blending of linear and nonlinear corrections provides the best extrapolation performance.

Unlike the Burgers case where fixed $\alpha=1$ caused catastrophic extrapolation failure, the cavity flow shows more graceful degradation with purely nonlinear corrections (C5). Nevertheless, C5 is still outperformed by C3, confirming that a balanced combination of linear and nonlinear corrections provides better generalization than either extreme alone.

\subsubsection{Effect of Network Width}

Experiments C6--C8 vary the network width while using  the optimal regularization from C3 ($\lambda_\alpha=10^{-5}$). This isolates the effect of network capacity on multifidelity correction performance.

The results reveal a capacity-accuracy trade-off. Narrow networks (C6, width 5) underfit the correction mapping, yielding the highest errors. Wider networks (C7, C8) progressively improve training accuracy but do not surpass the baseline architecture. The widest network (C8, width 15) achieves the best training fit among this group but at the cost of increased training time. Overall, the baseline architecture (C3) remains the best-performing configuration, indicating that moderate network capacity combined with appropriate regularization is more important than raw model size for this problem.

\begin{figure}[!htbp]
\centering
\includegraphics[width=0.95\textwidth]{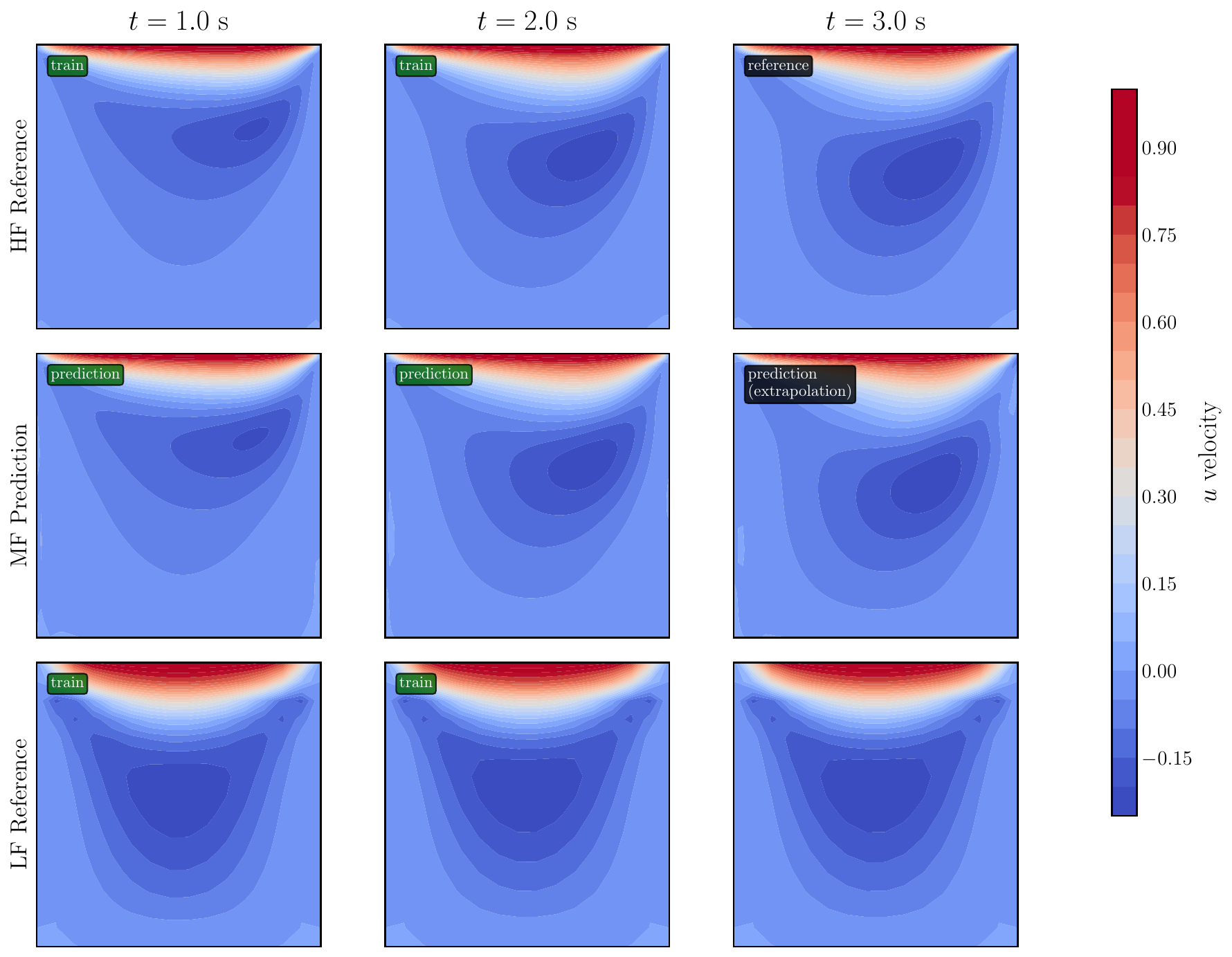}
\caption{Horizontal velocity $u(x,y)$ snapshots for lid-driven cavity flow at $t=1.0$ (training), $t=2.0$ (training boundary), and $t=3.0$ (extrapolation). Top: high-fidelity classical reference. Middle: multifidelity prediction (C3). Bottom: low-fidelity quantum (QLBM) solution.}
\label{fig:cavity_snapshots_u}
\end{figure}

\begin{figure}[!htbp]
\centering
\includegraphics[width=0.95\textwidth]{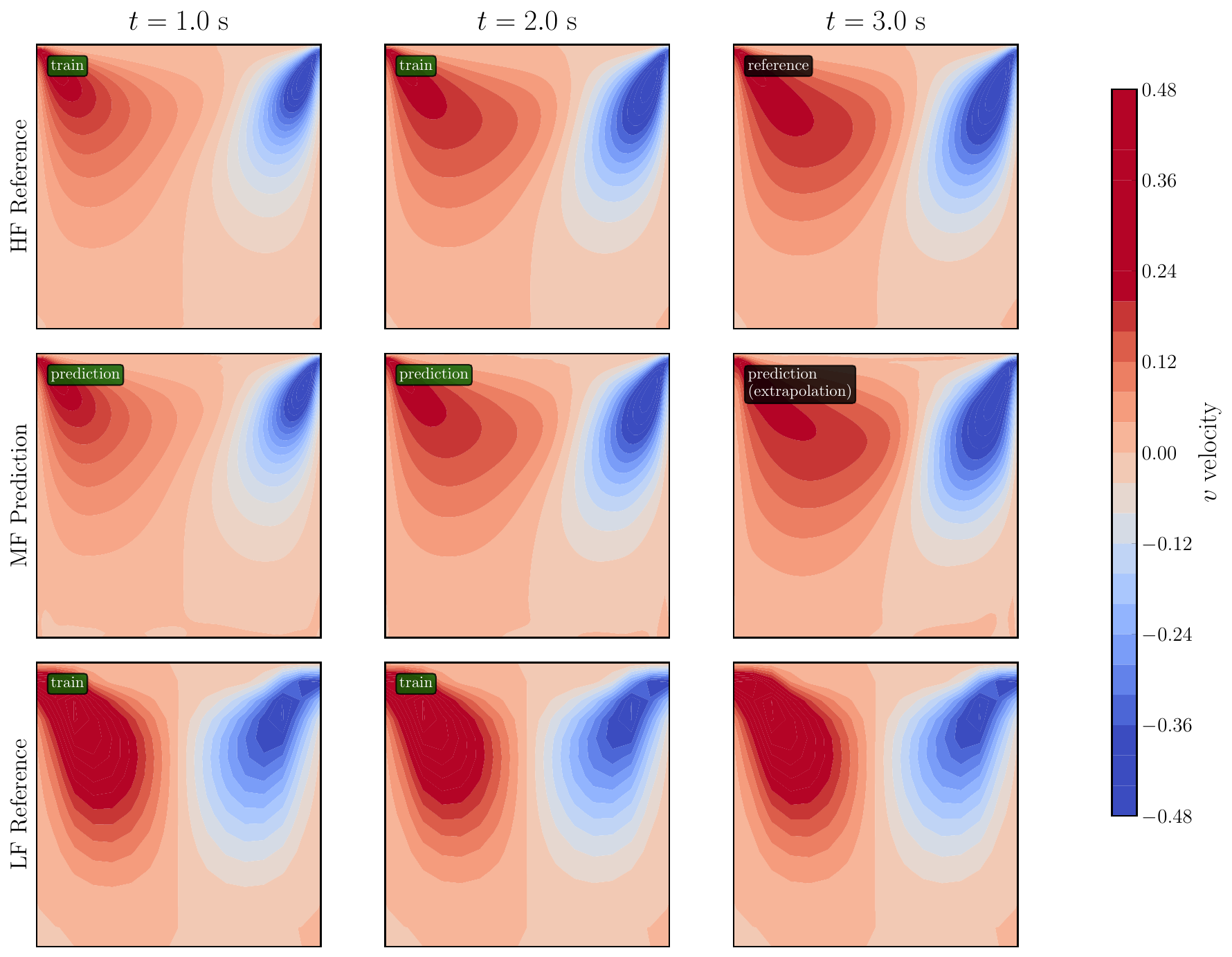}
\caption{Vertical velocity $v(x,y)$ snapshots for lid-driven cavity flow at $t=1.0$ (training), $t=2.0$ (training boundary), and $t=3.0$ (extrapolation). Top: high-fidelity classical reference. Middle: multifidelity prediction (C3). Bottom: low-fidelity quantum (QLBM) solution.}
\label{fig:cavity_snapshots_v}
\end{figure}

Figures~\ref{fig:cavity_snapshots_u} and~\ref{fig:cavity_snapshots_v} show velocity field snapshots at three time instances for the best-performing configuration (C3). The HF reference (top row) displays the characteristic lid-driven cavity flow pattern with a primary vortex centered in the upper portion of the cavity. The MF prediction (middle row) accurately captures both the spatial structure and temporal evolution of the flow, including in the extrapolation region ($t=3.0$). The LF quantum solution (bottom row) shows qualitatively correct behavior but with reduced spatial detail due to the coarse $16 \times 16$ grid.

\begin{figure}[!htbp]
\centering
\includegraphics[width=0.85\textwidth]{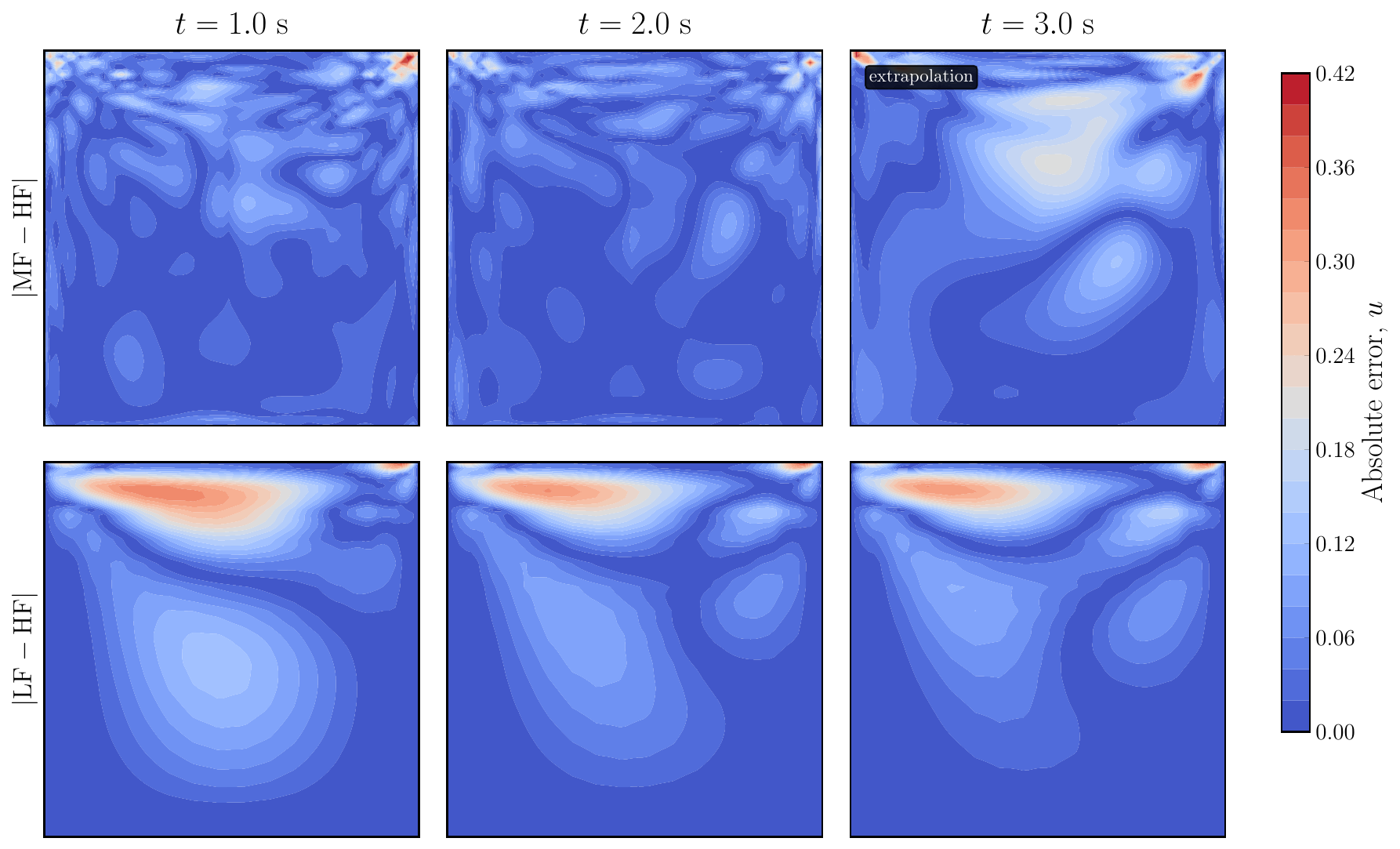}
\caption{Pointwise absolute errors for horizontal velocity $u$ in lid-driven cavity flow (C3). Top: multifidelity error $|u_{\text{MF}} - u_{\text{HF}}|$. Bottom: low-fidelity error $|u_{\text{LF}} - u_{\text{HF}}|$.}
\label{fig:cavity_error_u}
\end{figure}

\begin{figure}[!htbp]
\centering
\includegraphics[width=0.85\textwidth]{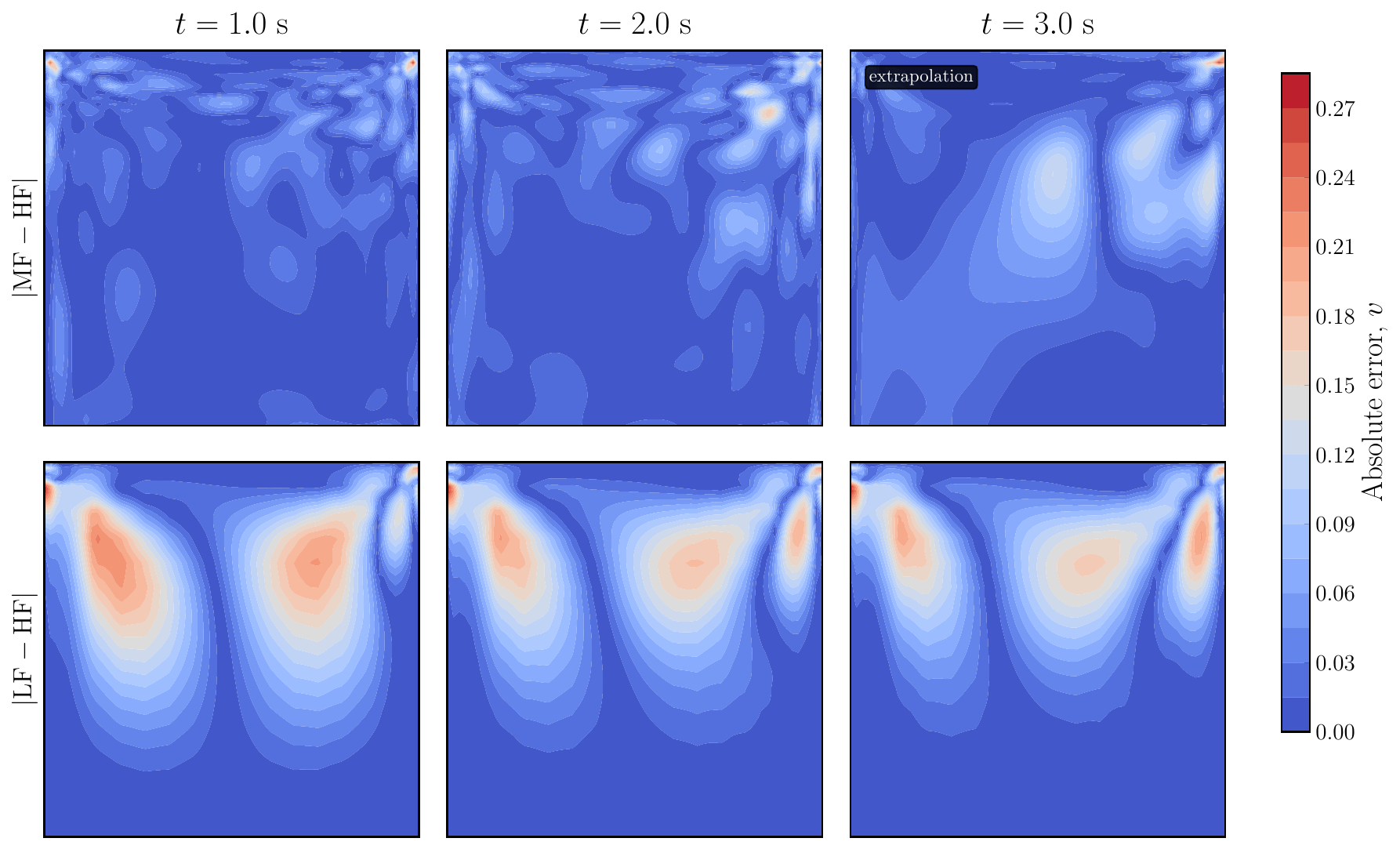}
\caption{Pointwise absolute errors for vertical velocity $v$ in lid-driven cavity flow (C3). Top: multifidelity error $|v_{\text{MF}} - v_{\text{HF}}|$. Bottom: low-fidelity error $|v_{\text{LF}} - v_{\text{HF}}|$.}
\label{fig:cavity_error_v}
\end{figure}

Figures~\ref{fig:cavity_error_u} and~\ref{fig:cavity_error_v} compare the pointwise absolute errors between MF and LF predictions relative to the HF reference. The MF correction substantially reduces errors throughout the domain for both velocity components, with the most significant improvements near the moving lid where velocity gradients are steepest.

\section{Conclusions}
\label{sec:conclusions}
We have demonstrated that multifidelity learning can effectively bridge quantum and classical solvers for nonlinear PDEs. Using quantum lattice Boltzmann methods as low-fidelity sources and classical finite-difference solvers as high-fidelity targets, the framework achieves substantial error reductions on both the viscous Burgers equation and lid-driven cavity flow.

This work advances the practical viability of near-term quantum computing for scientific simulation in two ways. First, it shows how to extract useful accuracy from quantum devices operating within current hardware constraints, providing immediate value rather than waiting for fault-tolerant systems. Second, it establishes a development pathway for quantum PDE algorithms: solvers can be designed and validated knowing that multifidelity correction can systematically bridge the gap to classical accuracy.

A key finding is that the learned corrections extrapolate beyond the high-fidelity training window, enabling accurate predictions over extended time horizons using only quantum data. This capability reduces dependence on expensive classical simulations and supports practical deployment scenarios. A future direction includes extending this framework by incorporating physics-informed constraints to further reduce high-fidelity data requirements.

\section{Code and data availability}
All code, trained models, and data required to replicate the examples presented in this paper will be released upon publication.

\section{Acknowledgements}
This project was completed with support from the U.S. Department of Energy, Advanced Scientific Computing Research program, under the "Uncertainty Quantification for Multifidelity Operator Learning (MOLUcQ)" project (Project No. 81739). The computational work was performed using PNNL Institutional Computing at Pacific Northwest National Laboratory. Pacific Northwest National Laboratory (PNNL) is a multi-program national laboratory operated for the U.S. Department of Energy (DOE) by Battelle Memorial Institute under Contract No. DE-AC05-76RL01830.

\bibliographystyle{my-elsarticle-num}
\bibliography{refs.bib}

\end{document}